\documentclass[runningheads]{llncs}

\usepackage{graphicx}
\usepackage{comment}
\usepackage{amsmath,amssymb} 
\usepackage{color}

\usepackage{multirow}
\usepackage[hidelinks]{hyperref}
\usepackage[font=small, labelfont=bf, labelsep=period]{caption}
\usepackage{subfig}
\usepackage{pifont}%
\newcommand{\putindex}[3]{\vtop{\hbox{\hspace{#3} $#1$}
            \hbox{\raise 6mm \hbox{$\scriptscriptstyle #2$}}}}

\newcommand{\gradx}[0]{\vtop{\hbox{\rm grad}
            \hbox{\raise 2.5mm \hbox{\rm \hspace{2mm} \footnotesize x}}}}

\newcommand{\grady}[0]{\vtop{\hbox{\rm grad}
            \hbox{\raise 2.5mm \hbox{\rm \hspace{2mm} \footnotesize y}}}}

\newcommand{\grad}[1]{\vtop{\hbox{\rm grad}
            \hbox{\raise 2.5mm \hbox{#1}}}}

\newcommand{\btb}{     \begin{tabbing}             }
\newcommand{\bte}{     \end{tabbing}               }

\newcommand{\eg}{e.g.}%
\newcommand{\etal}{et al.}%
\DeclareMathOperator*{\argmax}{argmax} 
\newcommand{\cmark}{\ding{51}}%
\newcommand{\xmark}{\ding{55}}%
\newdimen\owntablesep
\begin{document}
\pagestyle{headings}
\mainmatter
\def\ECCVSubNumber{3551}  

\title{Self-Supervised Monocular Depth Estimation:\\
Solving the Dynamic Object Problem\\ by Semantic Guidance} 

\titlerunning{Self-Supervised Depth Estimation With Semantic Guidance}
%
\author{Marvin Klingner \and Jan-Aike Term\"{o}hlen \and\\
Jonas Mikolajczyk \and Tim Fingscheidt}
\authorrunning{M. Klingner, J.-A. Term\"{o}hlen, J. Mikolajczyk, and T. Fingscheidt}
%
\institute{Technische Universit\"{a}t Braunschweig, Germany\\
\email{\{m.klingner, j.termoehlen, j.mikolajczyk, t.fingscheidt\}@tu-bs.de}
}
\maketitle

\begin{abstract}
	Self-supervised monocular depth estimation presents a powerful method to obtain 3D scene information from single camera images, which is trainable on arbitrary image sequences without requiring depth labels, \eg, from a LiDAR sensor. In this work we present a new self-supervised semantically-guided depth estimation (SGDepth) method to deal with moving dynamic-class (DC) objects, such as moving cars and pedestrians, which violate the static-world assumptions typically made during training of such models. Specifically, we propose (i) mutually beneficial cross-domain training of (supervised) semantic segmentation and self-supervised depth estimation with task-specific network heads, (ii) a semantic masking scheme providing guidance to prevent moving DC objects from contaminating the photometric loss, and (iii) a detection method for frames with non-moving DC objects, from which the depth of DC objects can be learned. We demonstrate the performance of our method on several benchmarks, in particular on the Eigen split, where we exceed all baselines without test-time refinement.
\end{abstract}

\begin{figure}[t]
	\centering	
	\includegraphics[width=0.65\linewidth]{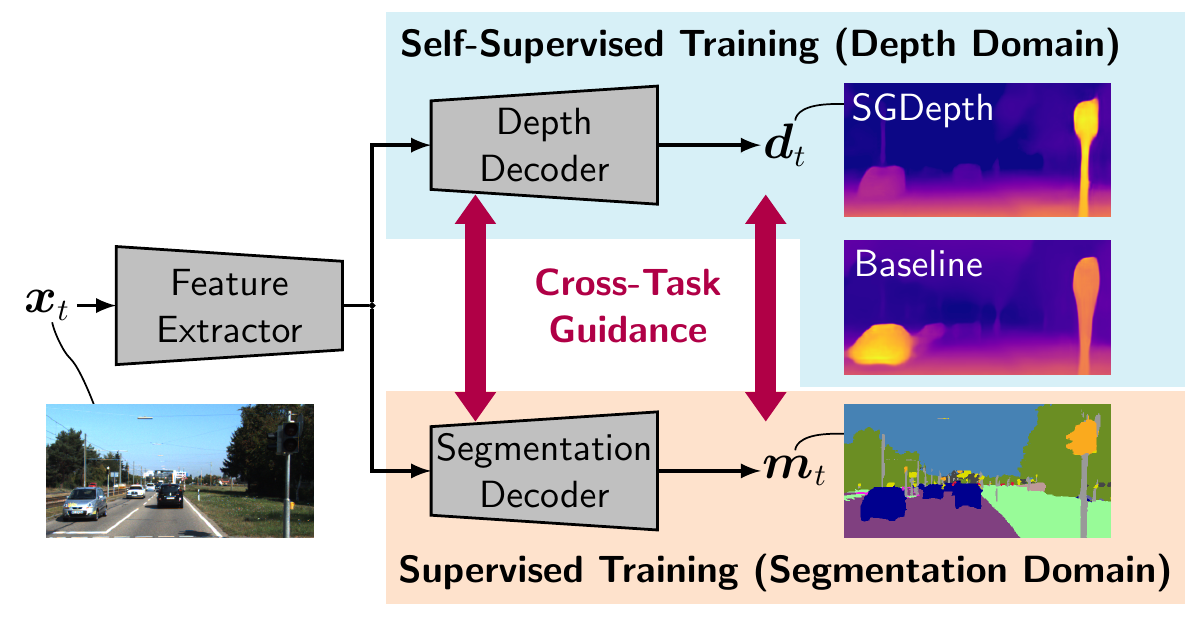}
	\caption{Overview over our framework for the combined prediction of semantic segmentation $\boldsymbol{m}_t$ and depth $\boldsymbol{d}_t$ from a single image $\boldsymbol{x}_t$ at time instant $t$. By combining \textbf{supervised training of semantic segmentation} in a source domain with \textbf{self-supervised training of depth} in a target domain, the segmentation masks guide the self-supervised monocular depth estimation inside the target domain.}
	\label{fig:general_overview}
\end{figure} 
\section{Introduction}
The accurate estimation of depth information from a scene is essential for applications requiring a 3D environment model such as autonomous driving or virtual reality. Therefore, a long-standing research field of computer vision is the prediction of depth maps from camera images. Classical model-based algorithms can predict depth from stereo images \cite{Hirschmuller2008} or from image sequences (videos) \cite{Akhter2009}, limited by the quality of the model. Deep learning enables the prediction of depth from single monocular images by supervision from LiDAR or RGB-D camera measurements \cite{Eigen2015,Eigen2014,Fu2018}. More recently, self-supervised approaches \cite{Garg2016,Godard2017} were introduced which solely rely on geometric image projection models and optimize the depth by minimizing photometric errors without the need of any labels. While these self-supervised monocular depth estimation approaches require only a single image as input during inference, they rely either on stereo images \cite{Garg2016}, or on sequential images from a video \cite{Zhou2017a} during training.
\par
For self-supervised monocular depth estimation from video data, the assumptions made during the geometric projections (which are required to calculate the photometric error) impose several problems: Firstly, occlusions can occur inducing artifacts in the photometric error. Secondly, consecutive more or less identical frames caused by a lack of ego-motion present a problem as without any movement between the frames no structure can be inferred. Thirdly, moving dynamic-class (DC) objects such as cars, trucks and pedestrians violate the static world assumption. Early approaches \cite{Mahjourian2018,Zhou2017a} did not address these problems. A current state-of-the-art approach by Godard \etal~\cite{Godard2019} approaches the first two problems by a minimum reprojection loss and an auto-masking technique, which we adopt (same as \cite{Casser2019,Guizilini2020a,Guizilini2020}). The third problem was left open in \cite{Casser2019,Godard2019,Guizilini2020a,Guizilini2020}.
\par
Starting to approach this dynamic object problem, we first need to identify dynamic-class (DC) objects pixel-wise by incorporating an image segmentation technique. For this purpose previous approaches either rely on pre-trained segmentation networks \cite{Casser2019,Casser2019a,Guizilini2020,Meng2019a}, which are not available for arbitrary datasets, or an implicit binary segmentation trained as part of the image projection model \cite{Luo2019a,Ranjan2019,Yang2018c}, thereby coupled and limited to the projection quality. Our solution is somewhat related to Chen \etal~\cite{Chen2019a}: We jointly optimize depth estimation and semantic segmentation, still keeping the depth estimation self-supervised by training the supervised semantic segmentation in a different domain. However, as \cite{Chen2019a} is limited to training on stereo images and proposes a unified decoder head for both tasks, we transfer it to the \textit{monocular} case and utilize gradient scaling described by \cite{Ganin2015} to enable cross-domain training with \textit{task-specific decoder heads}. This yields optimally learned task-specific weights inside the respective decoders and the \textit{possibility to generalize the concept} to even more tasks. 
\par
While we expect the depth estimation to take profit from sharper edges at object boundaries provided by semantic segmentation, the DC objects have to be handled once identified by the segmentation. In contrast to most other approaches \cite{Casser2019,Luo2019a,Meng2019a,Ranjan2019,Yang2018c}, we do not extend the image projection model to include DC objects, but simply exclude the pixels belonging to DC objects from the loss. However, this alone would lead to a poor performance, as the depth of DC objects would not be learned at all. Therefore, we propose a detection method for frames with \textit{non-moving} DC objects. From these frames the depth of (non-moving) DC objects can be learned with the normal (valid) image projection model, while in the other frames, the (moving) DC objects are excluded from the loss. Here, our approach presents a significantly simpler, yet powerful method to handle DC objects in self-supervised monocular depth estimation.\par
To sum up, our contribution to the field is threefold. Firstly, we generalize the mutually beneficial cross-domain training of self-supervised depth estimation and supervised semantic segmentation to a more general setting with task-specific network heads. Secondly, we introduce a solution to the dynamic object problem by using a novel semantically-masked photometric loss. Thirdly, we introduce a novel method of detecting \textit{moving} DC objects, which can then be excluded from the training loss computation, while \textit{non-moving} DC objects should still contribute. We demonstrate the effectiveness of our approach on the KITTI Eigen split, where we exceed all baselines without test-time refinement, as well as on two further KITTI benchmarks.\footnote{Code is available at \href{https://github.com/ifnspaml/SGDepth}{\url{https://github.com/ifnspaml/SGDepth}}.}

\section{Related Work}

Here, we give an overview about current methods for self-supervised depth estimation trained on sequential images. Afterwards we review how the dynamic object problem has been approached in multi-task learning settings.\par
\textbf{Depth Estimation:} 
Before the emergence of neural networks, stereo algorithms \cite{Hirschmuller2008,Sun2005} and structure from motion  \cite{Akhter2009,Ranftl2016} were used to infer depth from stereo image pairs or a series of images, respectively. Employing neural networks, Eigen \etal~\cite{Eigen2014} introduced the estimation of depth from a single image by training a network on sparse labels provided by LiDAR scans. Rapidly, the idea was further developed to improved architectures \cite{Eigen2015,Laina2017} and training techniques \cite{Kuznietsov2017,Liu2016}. Nowadays, many benchmarks \cite{Menze2015,Uhrig2017} in depth estimation are dominated by algorithms based on neural networks \cite{Fu2018,Zhang2019b}.\par
\textbf{Self-Supervised Depth Estimation:} 
More recently, self-supervised monocular depth estimation was proposed modeling depth as the geometric property of an image projection transformation between stereo image pairs \cite{Garg2016,Godard2017}, thereby optimizing a network based on the photometric error between the projected image and the actual image. Following, Zhou \etal~\cite{Zhou2017a} showed that it is possible to jointly optimize networks for the simultaneous prediction of depth and relative pose between two video frames. Since then this idea was complemented by improved loss functions \cite{Aleotti2018,Mahjourian2018}, specialized network architectures \cite{Guizilini2020a,Wang2018e,Zhou2019}, a hybrid approach utilizing video and stereo data \cite{Zhan2018}, and refinement strategies \cite{Casser2019,Casser2019a} to optimize the network or the prediction at test time. A state-of-the-art algorithm is presented by Godard \etal~\cite{Godard2019}, who propose a minimum reprojection loss to handle occlusions between different frames. For general image data it was proposed to additionally learn camera calibration parameters \cite{Gordon2019} or utilize additional depth labels from synthetic data \cite{Bozorgtabar2019}. Other approaches employ teacher-student learning \cite{Pilzer2019}, Generative Adversarial Networks (GANs) \cite{Aleotti2018,Kumar2018a,Pilzer2018}, proxy labels from traditional stereo algorithms \cite{Tosi2019}, or recurrent neural networks \cite{Wang2019,Zhang2019c}. However, as the used geometric projection relies on the assumption of a static world, current stand-alone algorithms for self-supervised monocular depth estimation trained on video are still not able to robustly handle moving DC objects.\par
\textbf{Multi-Task Learning:} 
Multi-task learning has shown improvements in many research fields, \eg, domain adaptation~\cite{Bolte2019a,Ochs2019,Zhao2019}, depth estimation \cite{Eigen2015,Ramirez2018,Zhang2019a} and semantic segmentation \cite{Kendall2018,Kirillov2019}. Yang \etal~\cite{Yang2018b} incorporated a semantic segmentation cue into the self-supervised depth estimation with stereo images as input, thus computing the cross-entropy loss between the predicted and respectively warped segmentation output scores of a network and corresponding ground truth labels. Chen \etal~\cite{Chen2019a} further develop this idea to single images at inference (still training on stereo images), and also compute losses between output scores of two stereo frames to supplement the photometric error, while supervising the semantic segmentation in another domain. However, their approach relies on a unified decoder structure for both tasks, whereas our approach generalizes cross-domain training to separate decoder heads and thereby better task-specific learned weights through the application of gradient scaling from \cite{Ganin2015}. Also we train on image sequences while \cite{Chen2019a} trains on stereo image pairs.\par
\textbf{Handling Dynamic-Class (DC) Objects:} 
In the following, we review other approaches to the dynamic object problem in self-supervised monocular depth estimation. Most existing works follow the classical way in considering optical flow, which can also be predicted in an unsupervised fashion \cite{Liu2019,Rhe2017}. By simultaneously predicting optical flow and depth, existing works impose losses for cross-task consistency \cite{Liu2019a,Luo2019a,Wang2019c,Yang2018c}, geometric constraints \cite{Chen2019b,Ranjan2019}, and modified reconstruction of the warped image \cite{Chen2019b,Yin2018}, all approaches extending the image projection model to moving DC objects. For example, Yang \etal~\cite{Yang2018c} predict a binary segmentation mask to identify moving DC objects and design loss functions for rigid and dynamic motion separately.\par
Our method is also related to approaches that rely on state-of-the-art segmentation techniques \cite{Casser2019,Casser2019a,Meng2019a,Tosi2020,Vijayanarasimhan2017}, which either give these as an additional input to the network \cite{Meng2019a} or use it to predict the relative pose between the same DC object in two consecutive frames \cite{Casser2019,Casser2019a,Vijayanarasimhan2017}, using this information to apply an additional separate rigid transformation for each DC object. Note, that all discussed related approaches add complexity to the geometric projection model in order to handle moving DC objects whereas our approach simply excludes DC objects from the loss, utilizing segmentation masks which are simultaneously and independently optimized by supervision from another domain. Finally, Li \etal~\cite{Li2019} propose to design a \textit{dataset} that consists solely of non-moving DC objects. Our method differs as we provide a \textit{detection method} for frames containing non-moving DC objects, from which the depth of DC objects can indeed be learned. 

\section{Method} 

\begin{figure*}[t]
	\centering
	\includegraphics[width=0.93\linewidth]{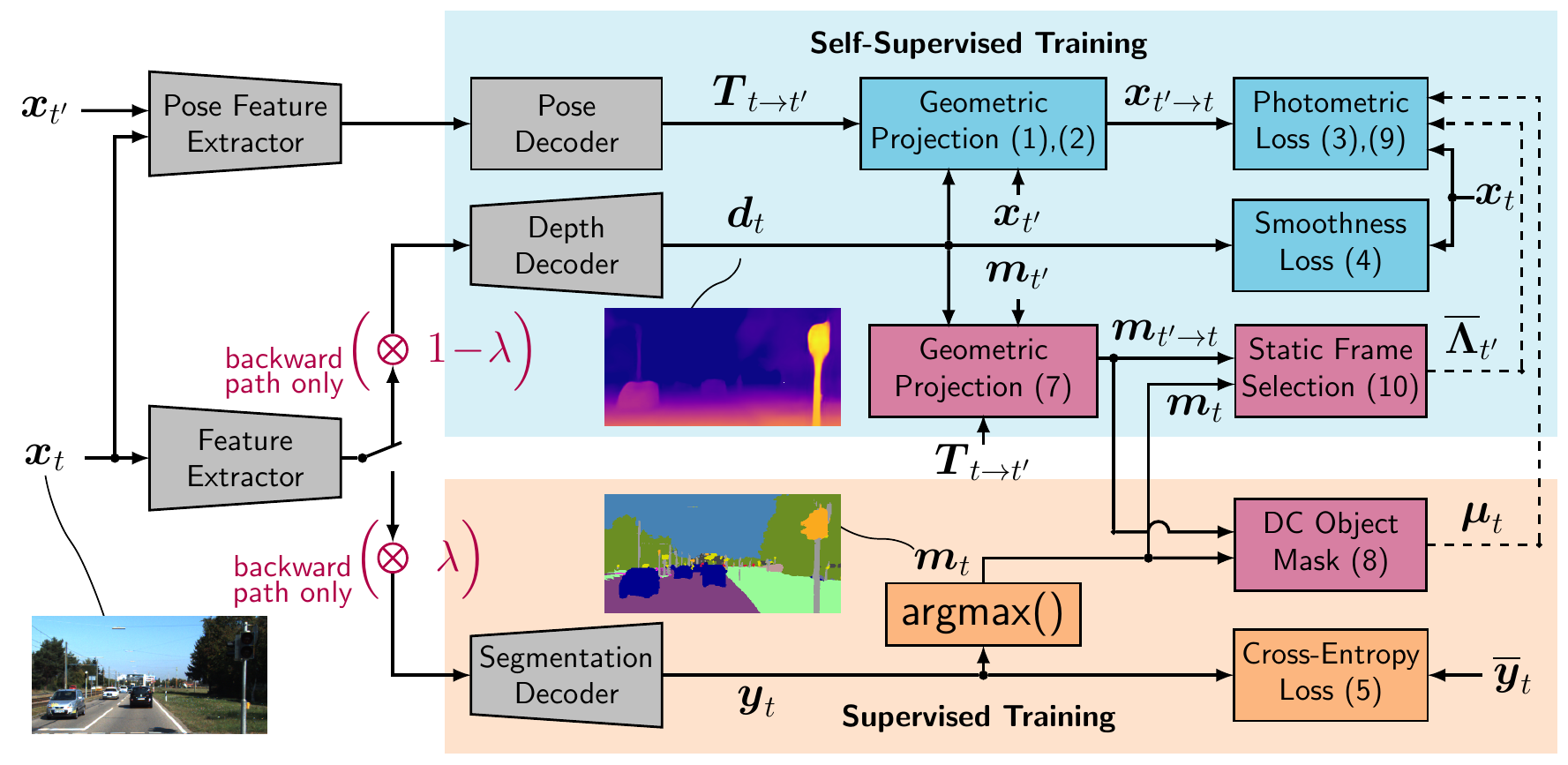}
	\caption{Overview over our \textbf{proposed framework} for \textbf{joint prediction of depth and semantic segmentation}. The grey blocks correspond to neural networks, the blue blocks correspond to the plain self-supervised depth estimation, the orange blocks correspond to the plain supervised semantic segmentation, and the red blocks correspond to semantic cross-task guidance between the two tasks. The numbers inside the blocks refer to the corresponding equations.}
	\label{fig:detailed_overview}
\end{figure*} 

In this part we will describe our framework (Fig.~\ref{fig:detailed_overview}). We first describe both predicted tasks independently. Afterwards we define our approach for solving the dynamic object problem by multi-task learning across domains and our novel semantic masking technique.

\subsection{Self-Supervised Monocular Depth Estimation}

Self-supervised monocular depth estimation defines the task of assigning depth values to camera image pixels without using any ground truth labels. Instead, the predicted depth is used as a geometric property to warp the frame at discrete time instance $t\! +\! 1$ to the previous frame at time $t$ with the photometric error between projected image and target image as the optimization objective.\par
\textbf{Inference Setting:} 
During inference, the neural network takes only a single RGB image $\boldsymbol{x}_t \in \mathbb{G}^{H\times W\times C}$ as input, where $\mathbb{G}$ is defined as the set of gray values $\mathbb{G} = \left\lbrace 0, 1, ..., 255 \right\rbrace$ of an image and $H$, $W$, and $C = 3$ define the height, the width, and the number of color channels, respectively. The output of the neural network is a dense depth map $\boldsymbol{d}_t \in \mathbb{D}^{H\times W}$ which assigns a depth to each pixel. The interval of possible depth values $\mathbb{D} = \left[d_{\mathrm{min}},d_{\mathrm{max}}\right]$ is defined by a lower bound $d_{\mathrm{min}}$ and an upper bound $d_{\mathrm{max}}$.\par
\textbf{Training Setting:} 
During training, the network utilizes preceding and succeeding frames $\boldsymbol{x}_{t'}$, with $t' \in \mathcal{T}' = \left\lbrace {t\!-\!1}, {t\!+\!1} \right\rbrace$, which are warped into the current frame at time $t$. This geometric transformation requires knowledge of the intrinsic camera parameter matrix $\boldsymbol{K} \in \mathbb{R}^{3\times 3}$, which we assume to be constant throughout one dataset and known in advance as in \cite{Godard2019}. Additionally, we require the prediction of the two relative poses $\boldsymbol{T}_{t\rightarrow t'}\in \mathit{SE}(3)$ between $\boldsymbol{x}_t$ and $\boldsymbol{x}_{t'}$, $t' \in \mathcal{T}'$, performed by the pose decoder in Fig.~\ref{fig:detailed_overview}. The special Euclidean group $\mathit{SE}(3)$ defines the set of all possible rotations and translations \cite{Szeliski2010}. While any such transformation is usually represented by a $4\times 4$ matrix $\boldsymbol{T}_{t\rightarrow t'}$, we follow \cite{Zhou2017a} in predicting only the six degrees of freedom. To predict the warped images $\boldsymbol{x}_{t'\rightarrow t}$, the image pixel coordinates $\boldsymbol{u}_{t} \in \mathcal{U}^{H\times W} =$ $\left\lbrace \left(h,w,1\right)^{\mathrm{T}} | h \in \left\lbrace 0,..., H\!\!-\!\!1\right\rbrace, w \in \left\lbrace 0,..., W\!\!\!-\!1\right\rbrace \right\rbrace$ are transformed to the pixel coordinate system at time $t'$, yielding coordinates $\boldsymbol{u}_{t\rightarrow t'}$, where $\left(\cdot\right)^{\mathrm{T}}$ denotes the vector transpose. Here, $\mathcal{U}^{H\times W}$ defines the set of pixel positions inside the image. For a single pixel coordinate $\boldsymbol{u}_{t, i} = \left(h_i,w_i,1\right)^{\mathrm{T}} \in \mathcal{U}$ with corresponding depth $d_{t,i}\in \mathcal{D}$ and  $i \in \mathcal{I} = \left\lbrace 1, ..., H W\right\rbrace$, the transformation can be written as \cite{Chen2019b}
\begin{equation}
	\boldsymbol{u}_{t\rightarrow t', i} = \underbrace{\left[\boldsymbol{K}| \mathbf{0} \right] \boldsymbol{T}_{t\rightarrow t'}}_{\substack{\text{transformation} \\ \text{to frame}\; t'}} \underbrace{\left[ \begin{pmatrix} d_{t,i} \boldsymbol{K}^{-1} \boldsymbol{u}_{t, i}\\ 1 \end{pmatrix} \right]}_{\substack{\text{projection to} \\ \text{3D point cloud}}},
	\label{eq:projection3d}
\end{equation}
with $\mathbf{0}$ being a three-dimensional zero vector. From right to left, the three parts can be interpreted as follows: First, the pixel with coordinate $\boldsymbol{u}_{t,i}\in\mathcal{U}$ is projected to the 3D space, afterwards the coordinate system is shifted by the relative pose $\boldsymbol{T}_{t\rightarrow t'}$, and finally the pixel is reprojected to the image at time $t'\in \mathcal{T}'$. We apply bilinear sampling $\mathrm{bil()}$ \cite{Jaderberg2015} to assign gray values to each pixel coordinate, as the projected coordinates $\boldsymbol{u}_{t\rightarrow t'}$ do not coincide with the pixel coordinates $\boldsymbol{u}_{t'}\in \mathcal{U}^{H\times W}$. In conclusion, the two warped images $\boldsymbol{x}_{t'\rightarrow t}$ are calculated as:
\begin{align}
	\boldsymbol{x}_{t'\rightarrow t} &= \operatorname{bil}\left(\boldsymbol{x}_{t'}, \boldsymbol{u}_{t\rightarrow t'}, \boldsymbol{u}_{t'}\right), t' \in \mathcal{T}'.
	\label{eq:bilinear_sampling}
\end{align}\par
\textbf{Minimum Reprojection Loss:} 
We follow common practice in choosing a mixture of absolute difference and structural similarity (SSIM) difference \cite{wang2004image} to compute the photometric loss $J_t^{\mathrm{ph}}$ between $\boldsymbol{x}_{t}$ and both $\boldsymbol{x}_{t'\rightarrow t}$, $t' \in \mathcal{T}'$, with a weighting factor $\alpha=0.85$ as in \cite{Mahjourian2018,Casser2019,Yin2018} . Adopting the \textit{per-pixel minimum} photometric loss \cite{Godard2019}, we get
\begin{align}
	J_t^{\mathrm{ph}} &= \left\langle \min_{t'\in\mathcal{T}'} \left( \frac{\alpha}{2}\left(\mathbf{1}-\boldsymbol{\mathrm{SSIM}}\left(\boldsymbol{x}_t, \boldsymbol{x}_{t'\rightarrow t}\right)\right) + \left(1-\alpha\right) |\boldsymbol{x}_t - \boldsymbol{x}_{t'\rightarrow t}| \right)\right\rangle,
	\label{eq:photometric_loss}
\end{align}
with $\mathbf{1}$ being a $H\times W$ matrix containing only ones, $\min\left(\cdot\right)$ of a matrix applying individually to each element (pixel position), $|\cdot|$ of a matrix delivering a matrix with its absolute elements, and $\left\langle \cdot \right\rangle$ representing the mean over all pixels. Note that $\boldsymbol{\mathrm{SSIM}}\left(\cdot\right) \in \mathbb{I}^{H\times W}, \mathbb{I} = \left[0, 1\right]$, is calculated on $3\times 3$ patches of the image.\par
\textbf{Smoothness Loss:} 
Encouraging pixels at nearby positions to have similar depths, we adapt the smoothness loss $J^{\mathrm{sm}}$ \cite{Godard2017,Godard2019} on the mean-normalized inverse depth $\overline{\boldsymbol{\rho}}_t\in \mathbb{R}^{H\times W}$, which is pixel-wise defined by $\rho_{t, i} = \frac{1}{d_{t, i}}$, and $\overline{\boldsymbol{\rho}}_t = \frac{\boldsymbol{\rho}_t}{\langle\boldsymbol{\rho}_t\rangle}$. The loss function is defined by
\begin{equation}
	J_t^{\mathrm{sm}} = \left\langle|\partial_h \overline{\boldsymbol{\rho}}_t|\exp\left(-|\partial_h \boldsymbol{x}_t|\right) + |\partial_w \overline{\boldsymbol{\rho}}_t|\exp\left(-|\partial_w \boldsymbol{x}_t|\right)\right\rangle,
	\label{eq:smoothness_loss}
\end{equation}
where $\partial_h$ and $\partial_w$ signify the one-dimensional difference quotient at each pixel position $\boldsymbol{u}_{t,i}\in \mathcal{U}$ with respect to the height and width direction of the image, respectively.\footnote{As an example, $\partial_w  \overline{\rho}_{t, i} = \overline{\rho}_{t, i+1} - \overline{\rho}_{t, i}$, under the condition that pixel index $i\!+\!1$ is in the same image row as $i$.} The smoothness loss allows large differences in depth only in regions with large differences between the gray values.

\subsection{Supervised Semantic Segmentation} 
The task of semantic segmentation is defined as assigning a label $m_{t, i} \in \mathcal{S}$ from a set of classes  $\mathcal{S} =  \left\lbrace1,2,...,S\right\rbrace$ to each pixel $\boldsymbol{x}_{t, i}$, which is achieved by a neural network that implements a non-linear mapping between the input image and output scores $\boldsymbol{y}_t\in \mathbb{I}^{H\times W \times S}$ for each pixel index $i$ and class $s\in \mathcal{S}$. Each element $p_{t, i,s}$ of the output scores $\boldsymbol{y}_t$ can be thought of as a posterior probability that the pixel $\boldsymbol{x}_{t,i}$ belongs to the class $s$. A segmentation mask $\boldsymbol{m}_t\in \mathcal{S}^{H\times W}$ can be obtained by computing $m_{t, i} = \argmax_{s \in \mathcal{S}} y_{t, i, s}$ and thus assigning a class to each pixel. The network is trained by imposing a weighted cross-entropy loss between the posterior probabilities of the network $\boldsymbol{y}_t$ and the ground truth labels $\overline{\boldsymbol{y}}_t$ with class weights $w_s$ \cite{Paszke2016}. Finally, again averaging over all pixels, the loss function for the image's posterior probabilities $\boldsymbol{y}_{t,s}\in \mathbb{I}^{H\times W}$ of class $s$ is defined as
\begin{equation}
J_t^{\mathrm{ce}} = -\left\langle\sum_{s \in\mathcal{S}} w_s \overline{\boldsymbol{y}}_{t,s} \odot \log\left(\boldsymbol{y}_{t,s}\right)\right\rangle,
\label{eq:crossentropy_loss}
\end{equation}
with $\log (\cdot)$ applied to each element of $\overline{\boldsymbol{y}}_{t,s}$, and $\odot$ standing for the element-wise multiplication between two matrices.

\begin{figure}[t]
	\centering
	\subfloat[][Image $\boldsymbol{x}_{t}$\label{fig:mask_analysis_a}]{\includegraphics[width=0.32\linewidth]{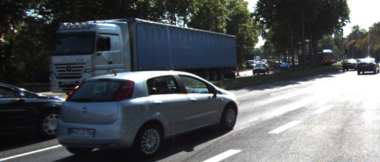}}\;\;%
	\subfloat[][Projected image $\boldsymbol{x}_{t\!-\!1\rightarrow t}$\label{fig:mask_analysis_c}]{\includegraphics[width=0.32\linewidth]{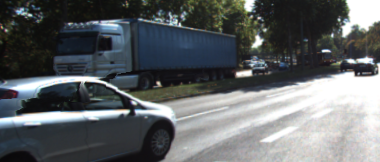}}\;\;%
	\subfloat[][Photometric error (\ref{eq:photometric_loss})\label{fig:mask_analysis_e}]{\includegraphics[width=0.32\linewidth]{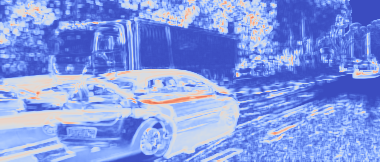}}\\[-0.25cm]
	\subfloat[][Segmentation $\boldsymbol{m}_{t}$\label{fig:mask_analysis_b}]{\includegraphics[width=0.32\linewidth]{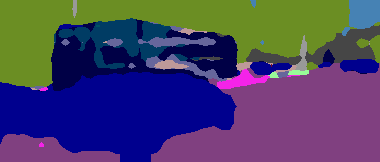}}\;\;%
	\subfloat[][Projected segmentation \phantom{} \quad\quad\quad\quad\quad $\boldsymbol{m}_{t\!-\!1\rightarrow t}$\!\!\label{fig:mask_analysis_d}]{\includegraphics[width=0.32\linewidth]{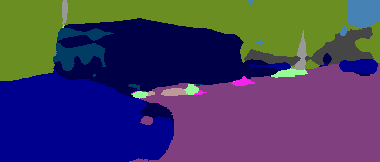}}\;\;%
	\subfloat[][DC object mask $\boldsymbol{\mu}_t$\label{fig:mask_analysis_f}]{\includegraphics[width=0.32\linewidth]{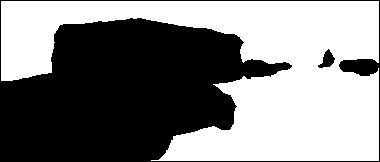}}\\%
	\caption{Example on how \textbf{moving DC objects can contaminate the photometric error}. Due to the movement of the car, the projected view in (b) is not valid, leading to unfavorable contributions for the photometric loss from (\ref{eq:photometric_loss}) as depicted in (c). This is addressed by masking the regions with potentially moving DC objects by calculating the DC object mask $\boldsymbol{\mu}_t$ (f) as in (\ref{eq:semantic_mask}) from the segmentation masks (d) and (e).}
	\label{fig:mask_analysis}
\end{figure} 

\subsection{Semantic Guidance} 

Now we describe our method to complement the depth estimation by a semantic masking strategy, which aims at resolving the problem of moving DC objects.\par
\textbf{Multi-Task Training Across Domains:}
We employ a single encoder with two decoder heads, one for the depth and one for the segmentation (see Fig.~\ref{fig:detailed_overview}). The decoder for the segmentation is trained in a source domain supervised by $\overline{\boldsymbol{y}}_{t,s}$, using (\ref{eq:crossentropy_loss}), while the decoder for the depth is trained in a target domain under self-supervision according to (\ref{eq:photometric_loss}) and (\ref{eq:smoothness_loss}). However, for mini-batches containing data from two domains, the question arises how to propagate the gradients from the separate decoders into the shared encoder. Other approaches weigh the loss functions by a factor \cite{Mahjourian2018,Zhou2017a} inducing the downside that the gradients inside the decoders are also scaled. Instead, we choose to follow \cite{Ganin2015} in scaling the gradients when they reach the encoder, see Fig.~\ref{fig:detailed_overview}. Let $\boldsymbol{g}^{\mathrm{depth}}$ and $\boldsymbol{g}^{\mathrm{seg}}$ be the gradients which are calculated according to the two decoders, then the total gradient $\boldsymbol{g}^{\mathrm{total}}$ propagated back to the encoder is calculated by:
\begin{equation}
	\boldsymbol{g}^{\mathrm{total}} = \left(1-\lambda\right) \boldsymbol{g}^{\mathrm{depth}} + \lambda \boldsymbol{g}^{\mathrm{seg}}.
	\label{eq:gradient_scaling}
\end{equation}\par
\textbf{Masking Out All DC Objects:} 
Motivated by the fact that moving DC objects contaminate the photometric error as shown in Fig.~\ref{fig:mask_analysis_e}, we want to mask out all DC objects that are present in the current frame $\boldsymbol{x}_t$ (Fig.~\ref{fig:mask_analysis_a}), as well as the wrongfully projected DC objects inside both projected frames $\boldsymbol{x}_{t'\rightarrow t}$, $t'\in \mathcal{T}'$, (Fig.~\ref{fig:mask_analysis_c}). Accordingly, we need to calculate both projected semantic masks $\boldsymbol{m}_{t'\rightarrow t}$, $t'\in \mathcal{T}'$, (Fig.~\ref{fig:mask_analysis_d}). To this end we apply nearest-neighbor sampling $\operatorname{near}\left(\cdot\right)$, where the interpolation strategy for the calculation of all pixels $\boldsymbol{x}_{t'\rightarrow t, i}$ from the bilinear sampling $\operatorname{bil}\left(\cdot\right)$ of \cite{Jaderberg2015} is replaced by assigning the value of the closest pixel inside of $\boldsymbol{m}_{t'}$ to the pixels of $m_{t'\rightarrow t, i}$, $i \in \mathcal{I}$. Consequently, the projected semantic mask can be calculated as:
\begin{equation}
\boldsymbol{m}_{t'\rightarrow t} = \operatorname{near}\left(\boldsymbol{m}_{t'}, \boldsymbol{u}_{t\rightarrow t'}, \boldsymbol{u}_{t'}\right).
\label{eq:semantic_warping}
\end{equation}\par
By defining DC object classes $\mathcal{S}_{\mathrm{DC}} \subset \mathcal{S}$, the DC object mask $\boldsymbol{\mu}_t\in \left\lbrace 0, 1\right\rbrace^{H\times W}$ is defined by its pixel elements:
\begin{equation}
\!\!\mu_{t, i} = 
\!\left\{
\begin{array}{l}
1 ,\; m_{t,i} \notin \mathcal{S}_{\mathrm{DC}}\; \land \; m_{t'\rightarrow t,i} \notin \mathcal{S}_{\mathrm{DC}} \; |\; t'\in \mathcal{T}' \\
0,\; \mathrm{else.} \\
\end{array}
\right.
\label{eq:semantic_mask}
\end{equation}
The mask contains $0$ at each pixel position $i$ belonging to a DC object in one of the three frames, and $1$ otherwise. Having obtained the DC object mask $\boldsymbol{\mu}_t$, we can define a semantically-masked photometric loss adapting (\ref{eq:photometric_loss}) 
\begin{align}
	J^{\mathrm{phm}}_{t} = &\left\langle \boldsymbol{\mu}_t\odot \min_{t'\in \mathcal{T}'} \left( \frac{\alpha}{2}\left(\mathbf{1}-\boldsymbol{\mathrm{SSIM}}\left(\boldsymbol{x}_t, \boldsymbol{x}_{t'\rightarrow t}\right)\right)+  \left(1-\alpha\right) |\boldsymbol{x}_t - \boldsymbol{x}_{t'\rightarrow t}| \right)\right\rangle,
	\label{eq:photometric_loss_masked}
\end{align}
which only considers non-DC pixels. We also consider the mask from the auto-masking technique \cite{Godard2019,Guizilini2020a,Guizilini2020}, which is omitted in (\ref{eq:photometric_loss}) and (\ref{eq:photometric_loss_masked}) for simplicity.

\begin{figure}[t]
	\centering	
	\includegraphics[width=0.9\linewidth]{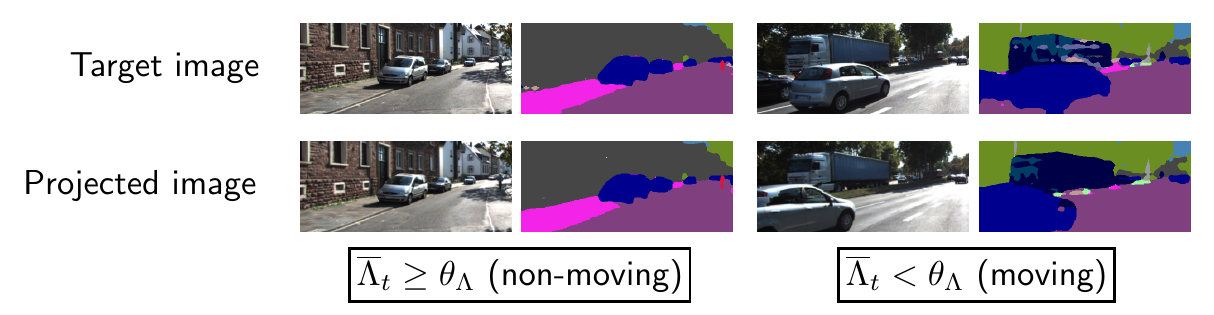}
	\caption{\textbf{Concept of the threshold} $\theta_{\Lambda}$ \textbf{in} (\ref{eq:total_loss}). For non-moving DC objects, target and projected segmentation mask are very similar (left), while they differ a lot for moving DC objects (right).}
	\label{fig:threshold_concept}
\end{figure} 

\textbf{Detecting Non-Moving DC Objects:} 
Inspired by \cite{Li2019}, we do not want to exclude the DC objects completely, instead we only learn from them, when they are not in motion. Accordingly, we need a measure to decide whether a DC object is in motion or not. The idea is based on the fact that if a DC object was observed to be in motion, the warped semantic mask in the target image $\boldsymbol{m}_{t'\rightarrow t}$ has a low consistency with the semantic mask $\boldsymbol{m}_{t}$ inside the target image, as shown in Fig.~\ref{fig:threshold_concept}. Accordingly, we can measure the intersection over union for dynamic object classes between $\boldsymbol{m}_{t'\rightarrow t}$ and $\boldsymbol{m}_{t}$ by:
\begin{align}
	\Lambda_{t, t'} = \frac{\sum_{i\in\mathcal{I}} \kappa_{t, t',i}}{\sum_{i\in\mathcal{I}} \nu_{t, t',i}}\;,\; \mathrm{with}\quad\label{eq:moving_object_iou}
	\kappa_{t,t',i} &= \left\{
\begin{array}{lll}
1 & , & m_{t,i} \in \mathcal{S}_{\mathrm{DC}}\;\; \land\;\; m_{t'\rightarrow t,i} \in \mathcal{S}_{\mathrm{DC}} \\
0 & , & \mathrm{else}, \\
\end{array}
\right.\\
 	\nu_{t,t',i} &= \left\{
\begin{array}{lll}
1 & , & m_{t,i} \in \mathcal{S}_{\mathrm{DC}}\;\; \lor\;\; m_{t'\rightarrow t,i} \in \mathcal{S}_{\mathrm{DC}} \\
0 & , & \mathrm{else}. \\
\end{array}
\right.\notag
\end{align}
The indicator $\Lambda_{t, t'}\in\left[ 0,1\right]$ signals perfect alignment and no moving DC objects if it equals $1$, while a value of $0$ indicates a high share of moving DC objects. If two frames at times $t'\in\mathcal{T}' = \left\lbrace t\!-\!1, t\!+\!1 \right\rbrace$ are considered, the mean value $\overline{\Lambda}_{t}$ of all $\Lambda_{t, t'}$ is to be taken. We define the threshold $\theta_{\Lambda}\in\left[ 0,1\right]$, above which an image is considered as static, see Fig.\ \ref{fig:threshold_concept}.\par
\textbf{Learning from Non-Moving DC Objects:} 
Having a measure that can indicate after each epoch whether an image is static or dynamic, we calculate $\overline{\Lambda}_{t}$ for each image of the dataset and choose the threshold $\theta_{\Lambda}$ such that a fraction $\epsilon \in \left[0, 1\right]$ of the images is trained without the semantically-masked photometric loss. The final loss is a combination of the photometric losses (\ref{eq:photometric_loss_masked}) and (\ref{eq:photometric_loss}), the smoothness loss (\ref{eq:smoothness_loss}), and the cross-entropy loss (\ref{eq:crossentropy_loss}), given by:
\begin{align}
J_{t}^{\mathrm{total}} = J_{t}^{\mathrm{ce}} + \beta J_{t}^{\mathrm{sm}} + \left\{
\begin{array}{lll}
J_{t}^{\mathrm{phm}} & , & \overline{\Lambda}_t < \theta_{\Lambda} \\
J_{t}^{\mathrm{ph}} & , & \mathrm{else}. \\
\end{array}
\right.
\label{eq:total_loss}
\end{align}
Note that $J_t^{\mathrm{ph}}$, $J_t^{\mathrm{phm}}$ and $J_t^{\mathrm{sm}}$ are computed only on images used for training of the depth, while $J_t^{\mathrm{ce}}$ is only computed in the domain of the segmentation, see Fig.~\ref{fig:detailed_overview}. Also note that in (\ref{eq:total_loss}) the segmentation and depth losses are not weighted against each other, as this weighting takes place in the midst of the backward pass guided by (\ref{eq:gradient_scaling}). 

\section{Experimental Setup}

In this section we describe the network topological aspects followed by the training details of our \texttt{PyTorch} \cite{Paszke2019} implementation. Afterwards, we describe the datasets and metrics used throughout our experimental evaluation.\par
\textbf{Network Topology}:
Our topology is based on \cite{Godard2019}, where an encoder-decoder architecture with skip connections is employed. To ensure comparability to existing work \cite{Casser2019,Guizilini2020,Godard2019,Wang2018e}, we choose an \texttt{Imagenet} \cite{Russakovsky2015} pretrained \texttt{ResNet18} encoder \cite{He2016}. The depth head has a sigmoid output $\sigma_{t,i}$, which is converted to a depth map by $\frac{1}{a \sigma_{t,i} + b}$, where $a$ and $b$ constrain the depth values to the range $\left[0.1, 100\right]$. For simplicity, the segmentation decoder uses the same architecture as the depth decoder, except for the last layer having $S$ feature maps, whose elements are converted to class probabilities by a softmax function. The pose network's architecture is the same as in \cite{Godard2019}.\par
\textbf{Training Aspects}:
For the training of the depth estimation, we resize all images to a resolution of $640\times 192$ ($416\times 128$ and $1280\times 384$ are also evaluated), if not mentioned otherwise, while for the semantic segmentation, the images are randomly cropped to the same resolution. We adopt the zero-mean normalization for the RGB images used during training of the \texttt{ResNet} encoder.  For input images we use augmentations including horizontal flipping, random brightness ($\pm 0.2$), contrast ($\pm 0.2$), saturation ($\pm 0.2$) and hue ($\pm 0.1$), while the photometric losses (\ref{eq:photometric_loss}, \ref{eq:photometric_loss_masked}) are calculated on images without color augmentations. We compute the loss on four scales as in \cite{Godard2019}.\par
We apply the gradient scaling from (\ref{eq:gradient_scaling}) at all connections between encoder and decoder with an empirically found optimal scale factor of $\lambda = 0.1$. The fraction $\epsilon$ of images, whose photometric loss is \textit{not masked} according to (\ref{eq:photometric_loss_masked}) is set to $0$ after 30 epochs and increased linearly, such that inside the last epoch the loss is calculated only according to (\ref{eq:photometric_loss}). This follows the idea that after removing the DC objects completely from the loss, the network is encouraged to learn from the frames with non-moving DC objects. We define DC object classes $\mathcal{S}_{\mathrm{DC}}$ as all classes belonging to the human and vehicle categories \cite{Cordts2016} (cf. Supp. A.2).\par
We train our models for 40 epochs with the Adam \cite{Kingma2015} optimizer and batch sizes of $12$ and $6$ for the single- and multi-task models, respectively. The batches from the two task-specific datasets are first concatenated, passed through the encoder, then disconnected and passed through the respective decoders. The learning rate is set to $10^{-4}$ and reduced to $10^{-5}$ after 30 epochs, as in \cite{Godard2019}. If we train only the depth estimation (with the architecture from \cite{Godard2019}), we dub it ``\textbf{SGDepth} only depth'', if both semantic segmentation and depth estimation are being trained according to our approach, we dub it ``\textbf{SGDepth} full''.\par
\textbf{Databases}:
We always utilize one dataset to train the semantic segmentation and another one for self-supervised training of the depth estimation of our SGDepth model. For training the semantic segmentation we utilize the \textit{Cityscapes} dataset \cite{Cordts2016} while at the same time we use different subsets of the KITTI dataset \cite{Geiger2013} for training the depth estimation. Similar to other state-of-the-art approaches we compare our depth estimation results by training and evaluating on the \textit{Eigen split} \cite{Eigen2014} of the KITTI dataset, following \cite{Zhou2017a} in removing static scenes from the training subset. We also train and evaluate on the single image depth prediction \textit{Benchmark split} from KITTI \cite{Uhrig2017}. To evaluate the joint prediction of depth and segmentation we utilize the \textit{KITTI split} defined by \cite{Godard2017} whose test set is the official training set of the KITTI Stereo 2015 dataset \cite{Menze2015}. The number of training images deviates slightly from the original definitions, as we need a preceding and a succeeding image \textit{to train} the depth estimation. The sizes of all data subsets are given in Table 4 of the appendix.\par
\textbf{Evaluation Metrics:} 
To evaluate the depth estimation we follow other works \cite{Mahjourian2018,Zhou2017a} in computing four error metrics between predicted and ground truth depth as defined in \cite{Eigen2014}, namely the absolute relative error (Abs Rel), the squared relative error (Sq Rel), the root mean squared error (RMSE), and the logarithmic root mean squared error (RMSE log). Additionally, we compute three accuracy metrics, which give the fraction $\delta$ of predicted depth values inside an image whose ratio and inverse ratio with the ground truth is below the thresholds $1.25$, $1.25^2$ and $1.25^3$. On the Benchmark split we evaluate using the scale-invariant logarithmic RMSE from \cite{Eigen2014} and the RMSE of the inverse depth (iRMSE). We follow \cite{Zhou2017a} by applying median scaling to the predicted depths. The semantic segmentation is evaluated using the mean intersection over union (mIoU) \cite{Everingham2015}, which is computed considering the classes as defined in \cite{Cordts2016}.

\section{Evaluation and Discussion}

In this section we start by a comparison to multiple state-of-the-art approaches, followed by an analysis how the single components of our method improve the results over our depth estimation and semantic segmentation baselines.

\subsection{Depth Evaluation w.r.t. the Baselines} 

The main evaluation is done on the \textit{Eigen split}, with the achieved results in Table~\ref{tab:comparison_state_of_the_art}. \textit{Our full SGDepth approach outperforms all comparable baselines,} where we compare to methods which use only image sequences as supervision on the target dataset (KITTI) and report results for the evaluation on single images at test-time. As we noted a high dependency of the results on the input resolution, we report our results on three resolutions. \textit{Note that at each resolution we outperform the baselines.} Furthermore, we provide results for our model, trained only with self-supervision for the depth estimation (SGDepth only depth), and show that the full SGDepth model is significantly better. Due to fairness, we do not compare against results with test-time refinement (\eg, in \cite{Casser2019}) or results employing a significantly larger network architecture (\eg, in \cite{Guizilini2020a}), as such techniques anyway can improve each of the methods further.\par
\begin{table*}[t!]
  \footnotesize
  \centering
  \setlength{\tabcolsep}{2pt}
  \caption{Evaluation of our new \textbf{self-supervised semantically-guided depth estimation} (SGDepth full) on the \textbf{KITTI Eigen split}. Baseline results are taken from the cited publications. For a fair comparison, we report results at 3 resolutions and \textbf{do not compare to methods with test-time refinement or significantly larger network architectures}. Additionally, we provide results for our model trained only for the depth estimation task (SGDepth only depth). CS indicates training of the depth estimation on Cityscapes, K training on the KITTI Eigen split, and (CS) training of the segmentation branch on Cityscapes. \textbf{Best results} at each resolution are written in \textbf{boldface} (the \texttt{ResNet50} model is out of competition).}
  \resizebox{\columnwidth}{!}{
  \begin{tabular}{|l|c|c|cccc|ccc|}
    \hline
  	 & & & \multicolumn{4}{c|}{Lower is better} & \multicolumn{3}{c|}{Higher is better}\\
  	 &&&&&&&&&\\[\owntablesep]
    \textbf{Method} & Resolution & Dataset & Abs Rel & Sq Rel & RMSE & RMSE log & $\delta < 1.25$ & $\delta < 1.25^2$ & $\delta < 1.25^3$\\
    &&&&&&&&&\\[\owntablesep]
    \hline        
    &&&&&&&&&\\[\owntablesep]
    Zhou \etal~\cite{Zhou2017a} & $416 \times 128$ & CS + K & $0.198$ & $1.836$ & $6.565$ & $0.275$ & $0.718$ & $0.901$ & $0.960$ \\
    Mahjourian \etal~\cite{Mahjourian2018} & $416 \times 128$ & CS + K & $0.159$ & $1.231$ & $5.912$ & $0.243$ & $0.784$ & $0.923$ & $0.970$ \\
    Yin and Shi~\cite{Yin2018} & $416 \times 128$ & CS + K & $0.153$ & $1.328$ & $5.737$ & $0.232$ & $0.802$ & $0.934$ & $0.972$ \\
    Wang \etal~\cite{Wang2018e} & $416 \times 128$ & CS + K & $0.148$ & $1.187$ & $5.583$ & $0.228$ & $0.810$ & $0.936$ & $0.975$ \\
    Casser \etal~\cite{Casser2019,Casser2019a} & $416 \times 128$ & K & $0.141$ & $1.026$ & $5.291$ & $0.215$ & $0.816$ & $0.945$ & $0.979$ \\
    Meng \etal~\cite{Meng2019a} & $416 \times 128$ & K & $0.139$ & $0.949$ & $5.227$ & $0.214$ & $0.818$ & $0.945$ & $\textbf{0.980}$ \\
    Godard \etal~\cite{Godard2018} & $416 \times 128$ & K & $0.128$ & $1.087$ & $5.171$ & $0.204$ & $0.855$ & $0.953$ & $0.978$ \\
    \textbf{SGDepth} only depth & $416 \times 128$ & K & $0.128$ & $1.003$ & $5.085$ & $0.206$ & $0.853$ & $0.951$ & $0.978$ \\
    \textbf{SGDepth} full & $416 \times 128$ & (CS) + K & $\textbf{0.121}$ & $\textbf{0.920}$ & $\textbf{4.935}$ & $\textbf{0.199}$ & $\textbf{0.863}$ & $\textbf{0.955}$ & $\textbf{0.980}$ \\
    \hline
    Guizilini \etal~\cite{Guizilini2020} & $640 \times 192$ & K & $0.117$ & $0.854$ & $4.714$ & $\textbf{0.191}$ & $0.873$ & $\textbf{0.963}$ & $\textbf{0.981}$ \\
    Godard \etal~\cite{Godard2018,Godard2019} & $640 \times 192$ & K & $0.115$ & $0.903$ & $4.863$ & $0.193$ & $0.877$ & $0.959$ & $\textbf{0.981}$ \\
    \textbf{SGDepth} only depth & $640 \times 192$ & K & $0.117$ & $0.907$ & $4.844$ & $0.196$ & $0.875$ & $0.958$ & $0.980$ \\  
    \textbf{SGDepth} full & $640 \times 192$ & (CS) + K & $\textbf{0.113}$ & $\textbf{0.835}$ & $\textbf{4.693}$ & $\textbf{0.191}$ & $\textbf{0.879}$ & $0.961$ & $\textbf{0.981}$ \\   
    \hline
    \textbf{SGDepth} full, \texttt{ResNet50} & $640 \times 192$ & (CS) + K & $0.112$ & $0.833$ & $4.688$ & $0.190$ & $0.884$ & $0.961$ & $0.981$ \\
    \hline
    Luo \etal~\cite{Luo2019a} & $832 \times 256$ & K & $0.141$ & $1.029$ & $5.350$ & $0.216$ & $0.816$ & $0.941$ & $0.976$ \\
    Ranjan \etal~\cite{Ranjan2019} & $832 \times 256$ & CS + K & $0.139$ & $1.032$ & $5.199$ & $0.213$ & $0.827$ & $0.943$ & $0.977$ \\
    Zhou \etal~\cite{Zhou2019} & $1248 \times 384$ & K & $0.121$ & $0.837$ & $4.945$ & $0.197$ & $0.853$ & $0.955$ & $\textbf{0.982}$ \\
    Godard \etal~\cite{Godard2018,Godard2019} & $1024 \times 320$ & K & $0.115$ & $0.882$ & $4.701$ & $0.190$ & $0.879$ & $0.961$ & $\textbf{0.982}$ \\  
    \textbf{SGDepth} only depth & $1280 \times 384$ & K & $0.113$ & $0.880$ & $4.695$ & $0.192$ & $0.884$ & $0.961$ & $0.981$\\ 
    \textbf{SGDepth} full & $1280 \times 384$ & (CS) + K & $\textbf{0.107}$ & $\textbf{0.768}$ & $\textbf{4.468}$ & $\textbf{0.186}$ & $\textbf{0.891}$ & $\textbf{0.963}$ & $\textbf{0.982}$ \\
    \hline
  \end{tabular}  
  } 
  \label{tab:comparison_state_of_the_art}
\end{table*}
\begin{figure*}[t!]
	\centering
	\includegraphics[width=1.0\linewidth]{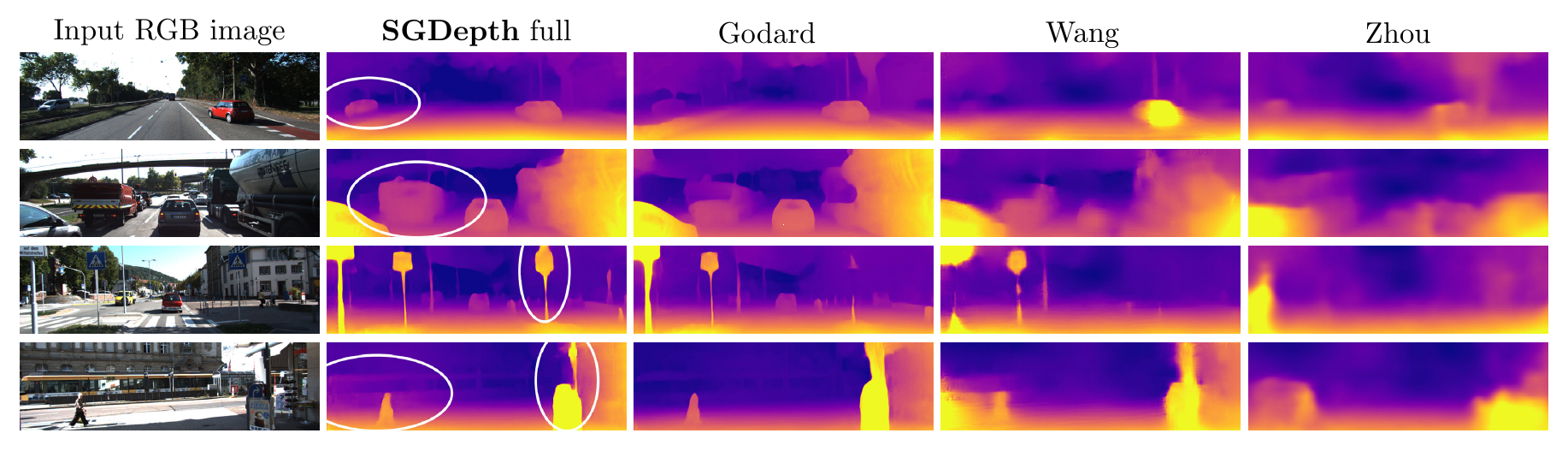}
	\put(-29.5,90){\scriptsize \cite{Zhou2017a}}
	\put(-98,90){\scriptsize \cite{Wang2018e}}
	\put(-165.5,90){\scriptsize \cite{Godard2019}}
	\caption{Qualitative examples of our full SGDepth method. Note: Boundaries of \textbf{DC objects are sharpened}, and in contrast to previous methods, \textbf{small objects} (\eg, traffic signs, third row) \textbf{are better detected/distinguished} by SGDepth full.}
	\label{fig:qualitative_results}
\end{figure*}
\begin{table}[t]
  \footnotesize
  \centering
  \setlength{\tabcolsep}{1.5pt}
  \caption{Results on the \textbf{KITTI depth prediction benchmark} (Benchmark\ split).}
  \resizebox{0.7\columnwidth}{!}{
  \begin{tabular}{|l|cccc|}
    \hline
  	 & \multicolumn{4}{c|}{Lower is better}\\
    \textbf{Method} & SILog & Abs Rel $\left[\%\right]$ & Sq Rel $\left[\%\right]$ & iRMSE\\
    \hline
    Fu \etal~\cite{Fu2018} (supervised) & $11.77$ & $2.23$ & $8.78$ & $12.98$ \\
    Ochs \etal~\cite{Ochs2019} (supervised) & $14.68$ & $3.90$ & $12.31$ & $15.96$ \\
    \textbf{SGDepth} full, \texttt{ResNet50} (self-supervised)& $15.30$ & $5.00$ & $13.29$ & $15.80$ \\        
    \textbf{SGDepth} full (self-supervised)& $15.49$ & $4.78$ & $13.33$ & $16.07$ \\  
    Goldman \etal~\cite{Goldman2019} (self-supervised)& $17.92$ & $6.88$ & $14.04$ & $17.62$ \\ 
    \hline
  \end{tabular}  
  }
  \label{tab:comparison_benchmark_depth}
\end{table}
Furthermore, in Table~\ref{tab:comparison_benchmark_depth} we provide results on the \textit{Benchmark split} for our full SGDepth model, which were computed on the KITTI online evaluation server. As we cannot use median scaling, we calculate a global scale factor on the validation split, which is applied before submitting the results for evaluation. Table~\ref{tab:comparison_benchmark_depth} shows that we outperform the only other listed self-supervised approach \cite{Goldman2019}, thereby reducing the gap to supervised methods \cite{Fu2018,Ochs2019}.\par
Qualitatively, we observe in Figure \ref{fig:qualitative_results} that the depth estimation has clearly shaped DC objects compared to the baselines. Furthermore, our SGDepth method is able to detect small objects such as traffic signs, where other methods fail.

\subsection{Ablation Studies} 

\begin{table*}[t!]
  \footnotesize
  \centering
  \setlength{\tabcolsep}{2pt}
  \caption{Evaluation of the \textbf{combined prediction of depth and semantic segmentation} on the \textbf{KITTI split} according to the standard protocol: We show how the single components of our approach improve the self-supervised depth estimation and how they compare to a stereo baseline. Note that in contrast to the stereo baseline all methods make use of median scaling. The values for mIoU scores on Cityscapes are obtained on the validation set. \textbf{Best results} overall and for monocular-trained methods are written in \textbf{boldface} (the \texttt{ResNet50} model is out of competition).}
  \resizebox{\columnwidth}{!}{
  \begin{tabular}{|l|cc|cccc|ccc|}
    \hline
      & \multicolumn{2}{c|}{Higher is better} & \multicolumn{4}{c|}{Lower is better} & \multicolumn{3}{c|}{Higher is better}\\
    &&&&&&&&&\\[\owntablesep]
    \textbf{Method} & $\mathrm{mIoU}_{\mathrm{K}}$ & $\mathrm{mIoU}_{\mathrm{CS}}$ & Abs Rel & Sq Rel & RMSE & RMSE log & $\delta < 1.25$ & $\delta < 1.25^2$ & $\delta < 1.25^3$\\
    &&&&&&&&&\\[\owntablesep]
    \hline   
    &&&&&&&&&\\[\owntablesep]
    Ramirez \etal~\cite{Ramirez2018} (stereo) & - & - & $0.143$ & $2.161$ & $6.526$ & $0.222$ & $ 0.850$ & $0.939$ & $0.972$ \\
    Chen \etal~\cite{Chen2019a} (stereo) & $37.7$ & $47.8$ & $0.102$ & $\textbf{0.890}$ & $\textbf{5.203}$ & $0.183$ & $0.863$ & $0.955$ & $0.984$ \\    
    Yang \etal~\cite{Yang2018c} (mono) & - & - & $0.131$ & $1.254$ & $6.117$ & $0.220$ & $0.826$ & $0.931$ & $0.973$ \\
    Liu \etal~\cite{Liu2019a} (mono) & - & - & $0.108$ & $1.020$ & $\textbf{5.528}$ & $0.195$ & $0.863$ & $0.948$ & $0.980$ \\
    \hline
    Or\u{s}i\'{c} \etal~\cite{Orsic2019} & - & $\textbf{75.5}$ & - & - & - & - & - & - & - \\
    \hline   
    \textbf{SGDepth} only segmentation & $43.1$ & $63.3$ & - & - & - & - & - & - & - \\
    \textbf{SGDepth} only depth & - & - & $0.108$ & $1.101$ & $6.379$ & $0.171$ & $0.878$ & $0.967$ & $0.988$ \\
    \hline   
     \textbf{SGDepth} add multi-task training & $42.6$ & $55.6$ & $0.105$ & $1.052$ & $6.298$ & $0.168$ & $0.882$ & $0.971$ & $0.990$ \\
    \textbf{SGDepth} add scaled gradients & $48.6$ & $67.7$ & $0.102$ & $1.023$ & $6.183$ & $0.164$ & $0.889$ & $0.972$ & $\textbf{0.991}$\\
    \textbf{SGDepth} add semantic mask & $48.3$ & $67.6$ & $0.106$ & $1.113$ & $6.337$ & $0.169$ & $0.884$ & $0.970$ & $0.989$ \\
    \textbf{SGDepth} add threshold & $\textbf{51.6}$ & $68.2$ & $0.099$ & $1.012$ & $6.120$ & $\textbf{0.160}$ & $0.894$ & $\textbf{0.973}$ & $0.990$ \\
    \hline
    \textbf{SGDepth} full & $50.1$ & $67.7$ & $\textbf{0.097}$ & $\textbf{0.983}$ & $6.173$ & $\textbf{0.160}$ & $\textbf{0.898}$ & $0.972$ & $0.990$ \\
    \textbf{SGDepth} full, \texttt{ResNet50} & $54.2$ & $70.7$ & $0.098$ & $0.940$ & $5.841$ & $0.156$ & $0.900$ & $0.976$ & $0.991$ \\
    \hline
  \end{tabular}
  }

  \label{tab:comparison_semantics}
\end{table*}
To show the effectiveness of our proposed improvements, we show results on the \textit{Kitti split} in Table \ref{tab:comparison_semantics}, starting from our baselines and individually adding our contributions up to our full method. Starting with our baselines trained only on the depth or the segmentation task, we observed depth estimation improvement when training both together in a multi-task fashion, where we simply add up the depth and segmentation losses from (\ref{eq:photometric_loss}), (\ref{eq:smoothness_loss}), and (\ref{eq:crossentropy_loss}). Note that the multi-task prediction of depth and segmentation is only done during training of our SGDepth models, while evaluation can be done separately for each task, inducing no additional complexity at test-time. Adding gradient scaling (\ref{eq:gradient_scaling}) improves on top, but particularly the semantic segmentation. In a first attempt to improve the depth estimation for DC objects, we masked out all DC objects that potentially contaminate the loss as described by (\ref{eq:semantic_mask}) and (\ref{eq:photometric_loss_masked}). Obviously, now the network does not get any objective on how to reconstruct the depth of DC objects, which leads to a decrease in performance. Therefore, we introduced the \textit{threshold} $\theta_{\Lambda}$ to learn the depth of DC objects from a fraction $\epsilon$ of images containing non-moving DC objects as described by (\ref{eq:moving_object_iou}) and (\ref{eq:total_loss}). For further improvement, we added a scheduling for the fraction $\epsilon$, to first learn the depth from the best samples, while afterwards allowing more and more ``noisy'' samples. \textit{Our final SGDepth model outperforms the Liu \etal~\cite{Liu2019a} mono approach in 6 out of 7 measures and even outperforms the stereo approach of Chen \etal~\cite{Chen2019a} in 5 out of 7 measures}.

\subsection{Semantics Evaluation} 

\begin{figure}[t]
	\raggedleft
	\subfloat[][Input RGB Image\label{fig:qualitative_a}]{
	\begin{minipage}[b][0.2cm][t]{.31\textwidth}
	\centering
	\includegraphics[width=1.0\linewidth]{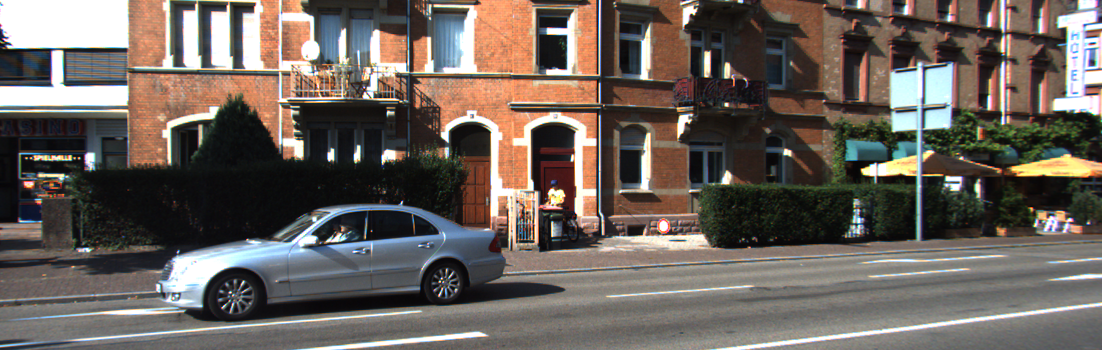}\\
	\textbf{(a)} Input RGB Image
	\end{minipage}
	}\;\;\!
	\subfloat[][\textbf{SGDepth} only depth\label{fig:qualitative_b}]{\includegraphics[width=0.31\linewidth]{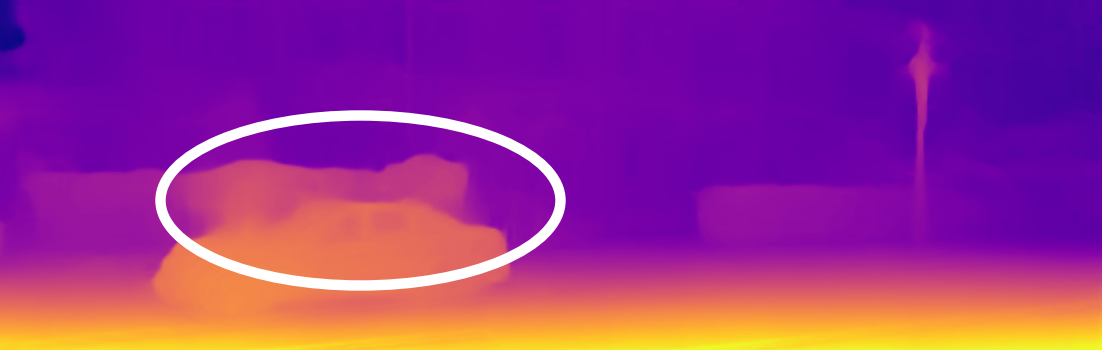}}\quad%
	\subfloat[][\textbf{SGDepth} full\label{fig:qualitative_c}]{\includegraphics[width=0.31\linewidth]{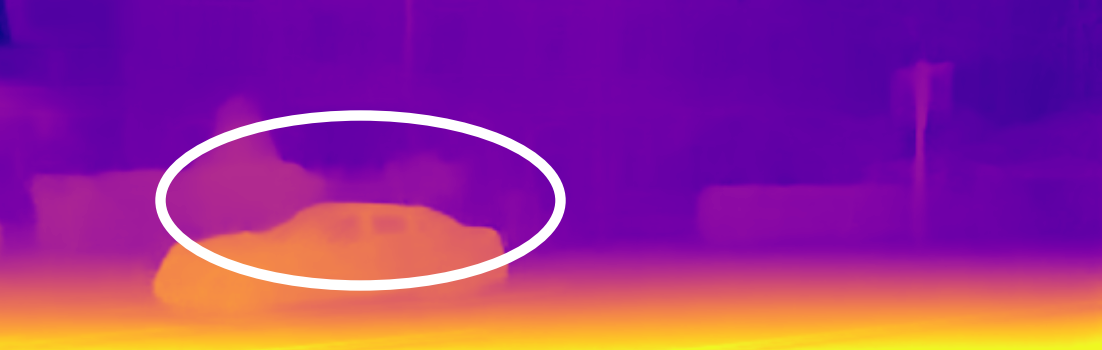}}\\[-0.3cm]
	\subfloat[][\textbf{SGDepth} only seg.\label{fig:qualitative_d}]{\includegraphics[width=0.31\linewidth]{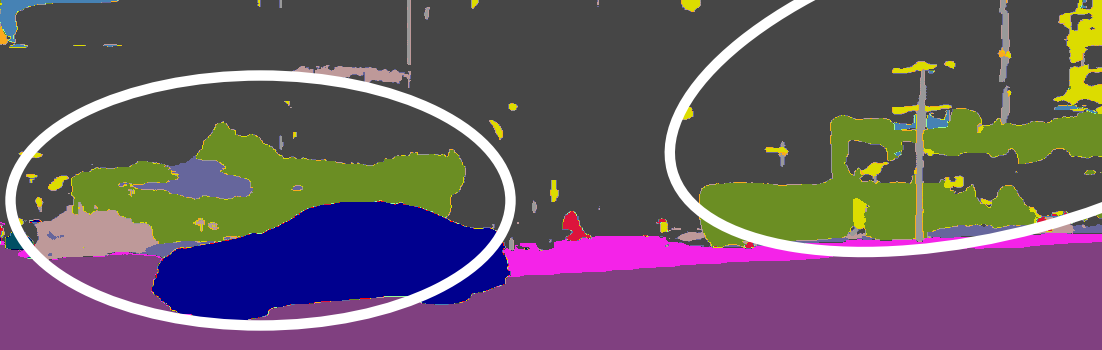}}\quad%
	\subfloat[][\textbf{SGDepth} full\label{fig:qualitative_e}]{\includegraphics[width=0.31\linewidth]{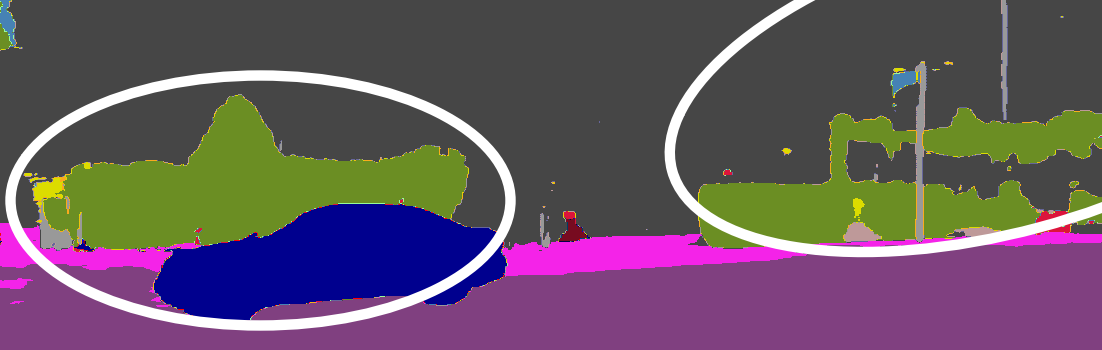}}
	\caption{Qualitative comparison between our \textbf{depth-} and \textbf{segmentation-only baselines} and our \textbf{full training approach}. Notice, how the depth boundaries are sharpened, while the artifacts inside the segmentation mask are reduced.}%
	\label{fig:depth_semantics_examples}
\end{figure}

The multi-task training of depth estimation and semantic segmentation not only achieves top results on the depth estimation, but also mutually improves the semantic segmentation in the source domain (Cityscapes), the semantic segmentation in the target domain (KITTI), and the depth estimation in the target domain (KITTI), as shown in Table \ref{tab:comparison_semantics}. We achieve a notable improvement from $43.1\%$ to $51.6\%$ on KITTI ($\mathrm{mIoU}_{\mathrm{K}}$) and from $63.3\%$ to $68.2\%$ on Cityscapes ($\mathrm{mIoU}_{\mathrm{CS}}$) for our best performing model on the segmentation task, denoted as ``\textbf{SGDepth} add threshold''. Our results are further improved when employing a larger \texttt{ResNet50} feature extractor. Additionally, Figure \ref{fig:depth_semantics_examples} shows that not only the depth boundaries of DC objects are sharpened but also the domain shift artifacts inside the semantic segmentation are significantly reduced.\par

\section{Conclusion}

In this work, we show how two tasks benefit from each other inside a multi-task cross-domain setting and develop a novel semantic masking technique to improve self-supervised monocular depth estimation for moving objects. We show superior performance on the KITTI Eigen split, exceeding all baselines without test-time refinement. We also demonstrate the effectiveness of each of our contributions on the KITTI split, where we outperform previous mono approaches in 6 out of 7 and even a stereo approach in 5 out of 7 measures.\par
Our approach is advantageous as long as the dataset used for training of our method contains some frames with non-moving dynamic-class (DC) objects belonging to the pre-defined semantic classes, \eg, parked vehicles, from which the depth of DC objects can be learned.

\bibliographystyle{splncs04}
\bibliography{ifn_spaml_bibliography}

\clearpage

\title{Supplementary} 

\author{\quad\\[-1cm]}
\institute{\quad}
\titlerunning{Self-Supervised Depth Estimation With Semantic Guidance}
%
%
\authorrunning{M. Klingner, J.-A. Termöhlen, J. Mikolajczyk, and T. Fingscheidt}

\maketitle

\appendix
\setcounter{table}{3}
\setcounter{figure}{6}
\setcounter{equation}{11}
\section{Experimental Setup}

In this part, we want to provide some additional details regarding our experimental setup, which allow a deeper understanding into our experimental setup.

\subsection{Detailed Dataset Overview}

In Table \ref{tab:comparison_dataset}, we give an overview over the number of images in our used data (sub)sets.  The Cityscapes dataset has 2,975 labeled training images on which we train the semantic segmentation part of our network. As we do not optimize our hyperparameters for the semantic segmentation and thereby do not use the validation set during training, our evaluation on this dataset is conducted on the official validation set containing 500 labeled images.\par
While we always train the segmentation part of our model on the Cityscapes dataset, the depth part of the network is trained on various splits of the KITTI dataset. The split of the KITTI dataset, which is most frequently used to compare depth estimation models, is the \textit{Eigen split} \cite{Eigen2014}, containing 697 images for testing. While the number of test images is constant throughout recent approaches, the number of training and validation images has been redefined by \cite{Zhou2017a} to exclude static scenes. We also compare our method on the \textit{Benchmark split} \cite{Uhrig2017}, which contains 500 test images with labels, which are only available on an evaluation server.\par
Finally, we train and evaluate on the \textit{KITTI split} \cite{Godard2017}, whose test set are the official 200 training images from the KITTI 2015 Stereo dataset \cite{Menze2015}. This test set has the advantage that it has available labels for both depth and semantic segmentation, which makes it suitable to observe the benefits of multi-task training for depth and semantic segmentation. While the Cityscapes validation set in principle also provides labels for both tasks, here the depth labels are obtained by a classical model-based algorithm, while the depth labels of the KITTI dataset are physical measurements from a LiDAR sensor and thereby better suited for evaluating a depth estimation model. Also, as our depth estimation training requires a preceding and a succeeding frame, the number of training images differs slightly from the original definition.

\begin{table}[t]
  \footnotesize
  \centering
  \setlength{\tabcolsep}{4pt}
  \caption{Overview over the used databases and available labels. Labels only available on a benchmark server are denoted by ``(\cmark)''.}
  \resizebox{0.6\columnwidth}{!}{
  \begin{tabular}{|lc|rcc|}
  \hline
  \multirow{2}{*}{Dataset} & \multirow{2}{*}{Subset} & \multirow{2}{*}{\# Images} & Depth & Segmentation\\
   & & & Labels & Labels\\
  \hline
  \multirow{3}{*}{Eigen split} & train & 21,880 & \cmark & \xmark\\
   & val & 4,187 & \cmark & \xmark\\
   & test & 697 & \cmark & \xmark\\
  \hline
  \multirow{3}{*}{Benchmark split} & train & 36,040 & \cmark & \xmark\\
   & val & 3,030 & \cmark & \xmark\\
   & test & 500 & (\cmark) & \xmark\\
  \hline
  \multirow{3}{*}{KITTI split} & train & 28,937 & \cmark & \xmark\\
   & val & 1,158 & \cmark & \xmark\\
   & test & 200 & \cmark & \cmark\\
  \hline
  \multirow{3}{*}{Cityscapes} & train & 2,975 & \cmark & \cmark\\
   & val & 500 & \cmark & \cmark\\
   & test & 1,525 & \cmark & (\cmark)\\
   \hline
  \end{tabular}
  }
  \label{tab:comparison_dataset}
\end{table}

\subsection{Definition of the DC Object Classes}
\label{sec:object_classes}

Also, we defined the DC object classes as all classes belonging to the human and vehicle categories inside the Cityscapes dataset \cite{Cordts2016} which contains in total 19 labeled classes. More specifically, that means that we consider the person, rider, car, truck, bus, train, motorcycle and bicycle class as DC object classes, as they are often observed as moving inside an image sequence. Opposed to that the classes road, sidewalk, building, wall, fence, pole, traffic light, traffic sign, vegetation, terrain and sky are considered as static, as they are usually not in motion.

\subsection{Evaluation Metrics}

In Section 4, we simply referred to previous approaches for the exact definition of the evaluation metrics. In this section, we provide the exact mathematical expressions which are used to evaluate the predicted depth maps $\boldsymbol{d}_t$ with available depth label $\overline{\boldsymbol{d}}_t$ as well as to evaluate the predicted segmentation maps $\boldsymbol{m}_t$ with regard to the ground truth label $\overline{\boldsymbol{m}}_t$. Note that the depth maps are evaluated using a sparse ground truth, where only the pixels with an available LiDAR measurement are considered during evaluation. Also, we apply median scaling to the predicted depth maps before evaluation to compensate the global scale ambiguity \cite{Zhou2017a}.\par
While the defintions of all metrics are equal for all data(sub)sets, there are two exceptions: On the Eigen test split we apply a crop defined by \cite{Eigen2014}, which is in accordance with previous approaches, while on the Benchmark test split we cannot apply median scaling, as the depth labels from the evaluation server are not freely available. Therefore, we determine the median over all image-wise scale factors on the validation set and use this value as a global scale factor for our predictions on the test set. Also note that the subsequent metrics (\ref{eq:abs_rel})-(\ref{eq:irmse}) are to be averaged over the respective test subset respectively.\par
The four error depth metrics used for evaluation on the Eigen and KITTI split are defined as:

\begin{align}
\text{Abs Rel} &= \frac{1}{HW}\sum_{i\in\mathcal{I}} \frac{|d_i-\overline{d}_i|}{\overline{d}_i}, \label{eq:abs_rel}\\
\text{Sq Rel} &= \frac{1}{HW}\sum_{i\in\mathcal{I}} \frac{\left(d_i-\overline{d}_i\right)^2}{\overline{d}_i},\\
\text{RMSE} &= \sqrt{\frac{1}{HW}\sum_{i\in\mathcal{I}} \left(d_i-\overline{d}_i\right)^2},\\
\text{RMSE log} &= \sqrt{\frac{1}{HW}\sum_{i\in\mathcal{I}} \left(\log d_i-\log \overline{d}_i\right)^2},
\end{align}

with $\mathcal{I}$ being the set of all pixels and $H$ and $W$ being the width and height of the image, respectively. The accuracy metrics are defined as follows:

\begin{align}
\text{The "$\delta < 1.25$" measure equals } & \frac{1}{HW}\sum_{i\in\mathcal{I}} \left[\max\left(\frac{d_i}{\overline{d}_i}, \frac{\overline{d}_i}{d_i}\right) < 1.25\right],\\
\text{the "$\delta < 1.25^2$" measure equals } & \frac{1}{HW}\sum_{i\in\mathcal{I}} \left[\max\left(\frac{d_i}{\overline{d}_i}, \frac{\overline{d}_i}{d_i}\right) < 1.25^2\right],\\
\text{the "$\delta < 1.25^3$" measure equals } & \frac{1}{HW}\sum_{i\in\mathcal{I}} \left[\max\left(\frac{d_i}{\overline{d}_i}, \frac{\overline{d}_i}{d_i}\right) < 1.25^3\right],\\
\end{align}

where $\left[\cdot\right]$ is defined as the Iverson bracket, which is $1$ if the condition inside the bracket is true, and $0$ if the condition is false. Furthermore on the benchmark split there are two more metrics, which are defined as

\begin{align}
\text{SILog} &= \frac{1}{HW}\sum_{i\in\mathcal{I}} \left(\log d_i-\log \overline{d}_i\right)^2 - \frac{1}{\left(HW\right)^2} \left(\sum_{i\in\mathcal{I}} \log d_i-\log \overline{d}_i\right)^2,\\
\text{iRMSE} &= \sqrt{\frac{1}{HW}\sum_{i\in\mathcal{I}} \left(\frac{1}{d_i}-\frac{1}{\overline{d}_i}\right)^2}.\label{eq:irmse}
\end{align}

Finally, we evaluate our semantic segmentation using the mean intersection over union (mIoU) metric, which is defined as:

\begin{align}
\mathrm{mIoU} = \frac{1}{S}\sum_{s\in\mathcal{S}}\frac{\mathrm{TP}_s}{\mathrm{TP}_s + \mathrm{FP}_s + \mathrm{FN}_s},
\end{align}

where $\mathcal{S} = \left\lbrace 1,2,...,S\right\rbrace$ is the set of all classes defined in the Cityscapes dataset as described in Section \ref{sec:object_classes}. Considering all labeled pixels for class $s\in\mathcal{S}$ in the predicted segmentation map $\boldsymbol{m}_t$, $\mathrm{TP}_s$ is the number of true positive predictions, $\mathrm{FP}_s$ is the number of false positive predictions, and $\mathrm{FN}_s$ is the number of false negative predictions. Note that while all depth metrics are computed image-wise and then averaged over all images inside the test set, for the mIoU calculation first $\mathrm{TP}_s$, $\mathrm{FP}_s$ and $\mathrm{FN}_s$ are summed up for all images of the test set and only afterwards the mIoU is calculated.

\section{Evaluation}

In this part, we give some additional examples of our proposed SGDepth method in comparison to several depth estimation baselines and also in comparison to our method trained without the semantic guidance (SGDepth only depth).

\begin{figure*}[t!]
	\centering
	\includegraphics[width=0.98\linewidth]{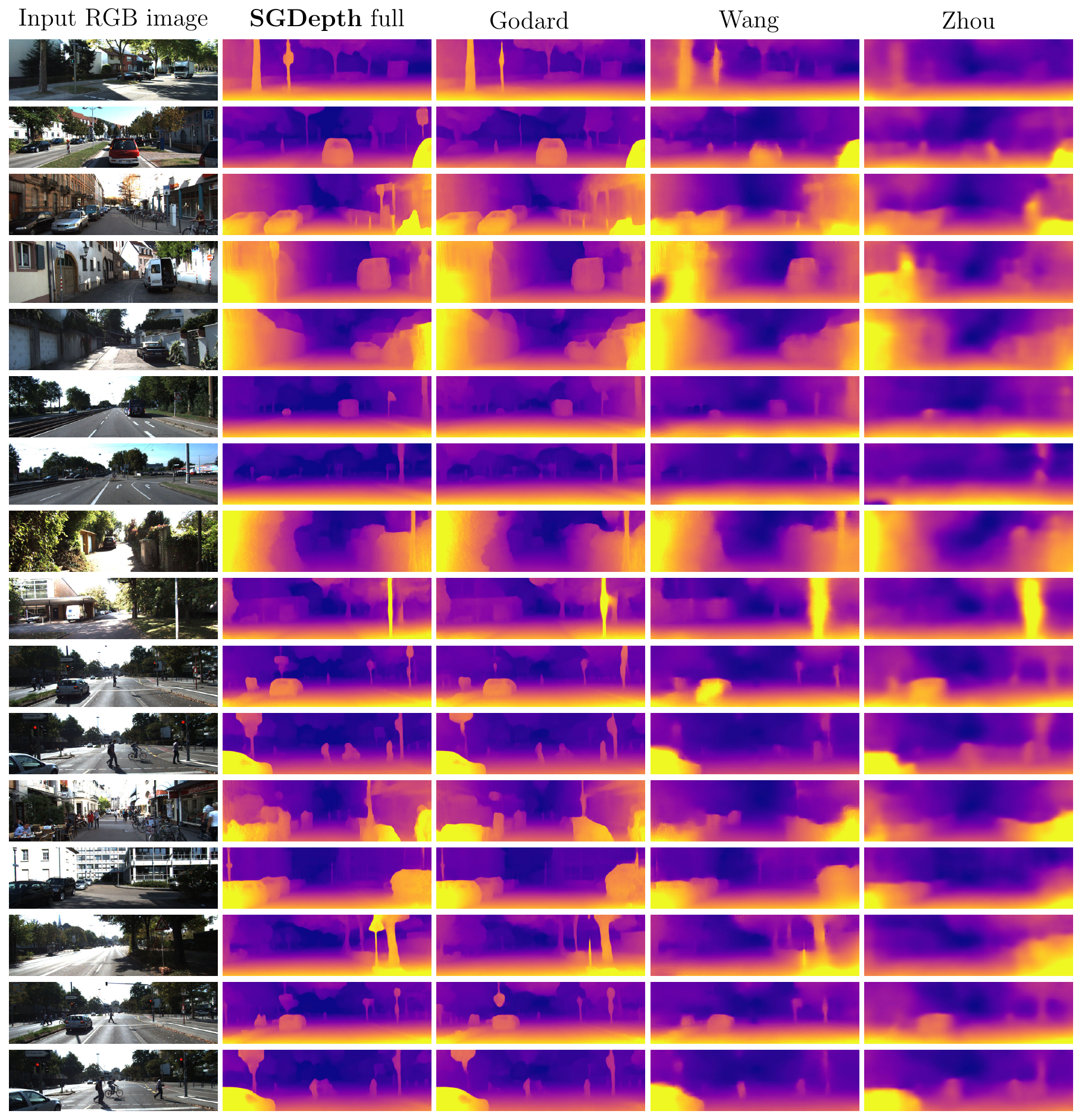}
	\put(-26,343.2){\scriptsize \cite{Zhou2017a}}
	\put(-94,343.2){\scriptsize \cite{Wang2018e}}
	\put(-160,343.2){\scriptsize \cite{Godard2019}}
	\caption{Additional examples of our proposed full SGDepth method in comparison to baseline methods. The figure is best viewed on screen and in color.}
	\label{fig:qualitative_results_1}
\end{figure*}

\begin{figure*}[t!]
	\centering
	\includegraphics[width=0.98\linewidth]{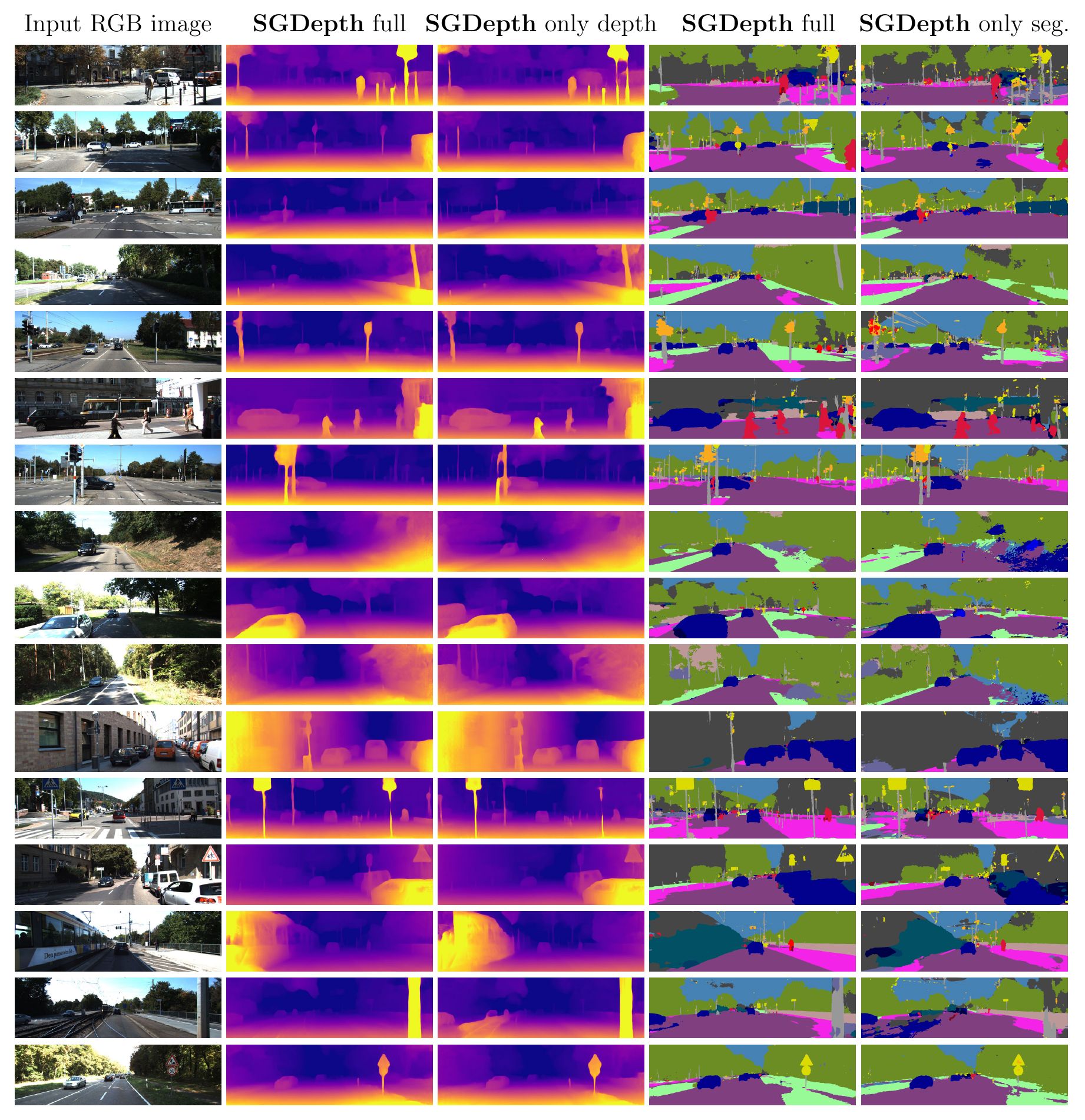}
	\caption{Additional examples on how the \textbf{full} SGDepth model compares to the models trained only on the \textbf{single tasks} of depth estimation and semantic segmentation, respectively. The figure is best viewed on screen and in color.}
	\label{fig:qualitative_results_2}
\end{figure*}

\subsection{Depth Comparison to Baselines}

In this section, we provide additional examples of the proposed SGDepth method, which we compare to results of the baseline approaches. All models were trained and tested on the Eigen splits \cite{Eigen2014} of the KITTI dataset \cite{Geiger2013}.\par
In the examples of Figure \ref{fig:qualitative_results_1} two things can be observed. Firstly, the depth predictions of our full SGDepth method are sharpened at object boundaries. This effect can be observed especially for small objects such as traffic lights or traffic signs as, \eg, in rows 1 and 3 from the top and row 3 from the bottom. This effect is mainly observed due to the joint training approach of depth estimation and semantic segmentation as thereby the encoder better learns to extract object boundaries, provided by the semantic segmentation, which in return guides the depth estimation to predict sharper edges at these boundaries. We also suspect that this effect is not even fully considered by the numerical evaluation as the ground truth depth labels only cover about the bottom two thirds of the image and many traffic signs and traffic lights are above this zone.\par
Secondly, our approach also allows for better learned depth of DC objects, as, \eg, in rows 1, 2 and 6 from the bottom, where especially the pedestrians and cyclists are more sharply visible inside the depth map. This is most likely due to our semantic masking technique, where the depth of DC objects is mainly learned from frames containing rather non-moving DC objects.

\subsection{Benefits of Multi-Task Training}

In this section we show additional examples of our SGDepth method for comparison with baselines trained only for the single tasks of depth estimation or semantic segmentation, respectively. The models were all trained and tested on the KITTI splits defined by \cite{Godard2017}.\par
The benefits of joint training for the depth estimation as discussed in the previous section also apply for our comparison to our own baseline, where one can clearly see the benefits of the semantic guidance for each single image inside Figure \ref{fig:qualitative_results_2}. However, also the semantic segmentation maps improve, compared to a semantic segmentation baseline (SGDepth only seg.), which was solely trained for the task of semantic segmentation (on the Cityscapes dataset). As stated in Section 5.3, we believe that this improvement on the KITTI dataset is due to the fact that through the self-supervised depth estimation, suitable features for the KITTI dataset are extracted, which bridge the domain shift between the Cityscapes and the KITTI dataset. This claim is also supported by the qualitative results, which appear clearly improved compared to the baseline.

\subsection{KITTI Eigen Split Ablation}

\begin{table*}[t!]
  \footnotesize
  \centering
  \setlength{\tabcolsep}{2pt}
  \caption{\textbf{Ablation study} of different models on the \textbf{KITTI Eigen split}. CS indicates training of the depth estimation on Cityscapes, K training on the KITTI Eigen split, and (CS) training of the segmentation branch on Cityscapes. \textbf{Best results} at each resolution are written in \textbf{boldface}.}
  \resizebox{\columnwidth}{!}{
  \begin{tabular}{|l|c|c|cccc|ccc|}
    \hline
  	 & & & \multicolumn{4}{c|}{Lower is better} & \multicolumn{3}{c|}{Higher is better}\\
  	 &&&&&&&&&\\[\owntablesep]
    \textbf{Method} & Resolution & Dataset & Abs Rel & Sq Rel & RMSE & RMSE log & $\delta < 1.25$ & $\delta < 1.25^2$ & $\delta < 1.25^3$\\
    &&&&&&&&&\\[\owntablesep]
    \hline        
    &&&&&&&&&\\[\owntablesep]
    \textbf{SGDepth} only depth & $640 \times 192$ & K & $0.117$ & $0.907$ & $4.844$ & $0.196$ & $0.875$ & $0.958$ & $0.980$ \\
    &&&&&&&&&\\[\owntablesep]
    \hline        
    &&&&&&&&&\\[\owntablesep]
     \textbf{SGDepth} add multi-task training & $640 \times 192$ & (CS) + K &   $0.117$ & $0.918$ & $4.777$ & $0.193$ & $0.872$ & $0.960$ & $\textbf{0.982}$ \\
    \textbf{SGDepth} add scaled gradients & $640 \times 192$ & (CS) + K & $\textbf{0.113}$ & $\textbf{0.817}$ & $\textbf{4.671}$ & $0.191$ & $0.877$ & $\textbf{0.961}$ & $\textbf{0.982}$ \\
    \textbf{SGDepth} add semantic mask & $640 \times 192$ & (CS) + K & $0.116$ & $0.917$ & $4.726$ & $\textbf{0.189}$ & $0.874$ & $\textbf{0.961}$ & $\textbf{0.982}$ \\
    \textbf{SGDepth} add threshold & $640 \times 192$ & (CS) + K & $\textbf{0.113}$ & $0.861$ & $4.724$ & $0.191$ & $\textbf{0.879}$ & $0.960$ & $0.981$ \\
    &&&&&&&&&\\[\owntablesep]
    \hline        
    &&&&&&&&&\\[\owntablesep]
    \textbf{SGDepth} full & $640 \times 192$ & (CS) + K & $\textbf{0.113}$ & $0.835$ & $4.693$ & $0.191$ & $\textbf{0.879}$ & $\textbf{0.961}$ & $0.981$ \\
    \hline
  \end{tabular}  
  } 
  \label{tab:comparison_eigen_ablation}
\end{table*}

We trained and optimized the parameters of our different model variants on the KITTI split \cite{Godard2017} to observe the resulting performance on both tasks, depth and semantic segmentation. In the end we only evaluated the final obtained models on the Eigen split benchmark. However, for completeness we also provide the same ablation experiments as executed on the KITTI split on the Eigen split in Table~\ref{tab:comparison_eigen_ablation}. We observe that all multi-task models outperform the single-task baseline (SGDepth only depth) and that our final model (SGDepth full) is best in the important metrics Abs. Rel. and $\delta < 1.25$, as has been observed on the KITTI split as well. \textit{Thereby, our ablation on the Eigen split confirms our ablation experiments on the KITTI split}.

\subsection{Pose Evaluation}

\begin{table}[t]
  \footnotesize
  \centering
  \caption{Pose estimation results on the KITTI odometry dataset sequences 9 and 10.}
  \setlength{\tabcolsep}{4pt}
  \resizebox{0.6\columnwidth}{!}{
  \begin{tabular}{|l|c|c|c|}
  \hline
  Method & Sequence 9 & Sequence 10 & \# frames\\
  \hline
  Zhou \etal~\cite{Zhou2017a} & $0.021 \pm 0.017$ & $0.020 \pm 0.015$ & 5 \\
  Godard \etal~\cite{Godard2018} & $0.017\pm 0.008$ & $0.015 \pm 0.010$ & 2 \\
  Luo \etal~\cite{Luo2019a} & $0.013 \pm 0.007$ & $\textbf{0.012}\pm \textbf{0.008}$ & 3 \\
  Ranjan \etal~\cite{Ranjan2019} & $\textbf{0.012} \pm \textbf{0.007}$ & $\textbf{0.012} \pm \textbf{0.008}$ & 5 \\  
  \hline
  \textbf{SGDepth} only depth & $0.017 \pm 0.009$ & $0.014 \pm 0.010$ & 2 \\
  \textbf{SGDepth} full & $0.019 \pm 0.010$ & $0.016 \pm 0.010$ & 2 \\
  \hline
  \end{tabular}
  }
  \label{tab:comparison_pose}
\end{table}

Although the focus of our work is on depth estimation, we also provide results of our pose estimation network evaluated with the same strategy as introduced in \cite{Godard2019,Zhou2017a}. We trained on the sequences 0 to 8 of the KITTI odometry dataset and evaluated our models on the sequences 9 and 10 with the results compared to baselines shown in Table \ref{tab:comparison_pose}. Interestingly, the joint training of depth and semantic segmentation seems to have a negative effect on the pose estimation, whose optimization through multi-task learning could be subject to future works. Nevertheless we achieve competitive results compared to the baselines \cite{Luo2019a,Ranjan2019}, in particular when considering that most of them use more than 2 input images for pose estimation at test time.

\end{document}